\documentclass[10pt,twocolumn,letterpaper]{article}

\usepackage{iccv}
\usepackage{times}
\usepackage{epsfig}
\usepackage{graphicx}
\usepackage{amsmath}
\usepackage{amssymb}
\usepackage{microtype}

\usepackage{booktabs}
\usepackage{multirow}
\usepackage[ruled]{algorithm2e}
\usepackage{algpseudocode}
\usepackage{bbm}
\usepackage{adjustbox}

\usepackage[pagebackref=true,breaklinks=true,letterpaper=true,colorlinks,bookmarks=false]{hyperref}

\newcommand{\rom}[1]{\uppercase\expandafter{\romannumeral #1\relax}}
\usepackage[capitalize]{cleveref}
\crefname{section}{Sec.}{Secs.}
\Crefname{section}{Section}{Sections}
\Crefname{table}{Table}{Tables}
\crefname{table}{Tab.}{Tabs.}
\newcommand{\squeezeup}{\vspace{-4mm}}

\iccvfinalcopy % *** Uncomment this line for the final submission

\def\iccvPaperID{7068} % *** Enter the ICCV Paper ID here

\ificcvfinal\fi

\begin{document}

%%%%%%%%% TITLE
\title{VADER: Video Alignment Differencing and Retrieval}

% \and

\author{Alexander Black$^1$
        \quad Simon Jenni$^2$
        \quad Tu Bui$^1$
        \quad Md. Mehrab Tanjim$^3$
        \quad Stefano Petrangeli$^2$\\
        \quad Ritwik Sinha$^2$
        \quad Viswanathan Swaminathan$^2$
        \quad John Collomosse$^{1,2}$\\
        $^1$CVSSP, University of Surrey
        \quad $^2$Adobe Research\\
        \quad $^3$University of California, San Diego\\
        \tt\small {\{alex.black,t.v.bui\}@surrey.ac.uk \quad\{jenni,petrange,risinha,vishy,collomos\}@adobe.com}
       }

\twocolumn[{%
\renewcommand\twocolumn[1][]{#1}%
\maketitle
\vspace{-25pt}
\begin{center}
    \centering
    \includegraphics[width=0.9\linewidth]{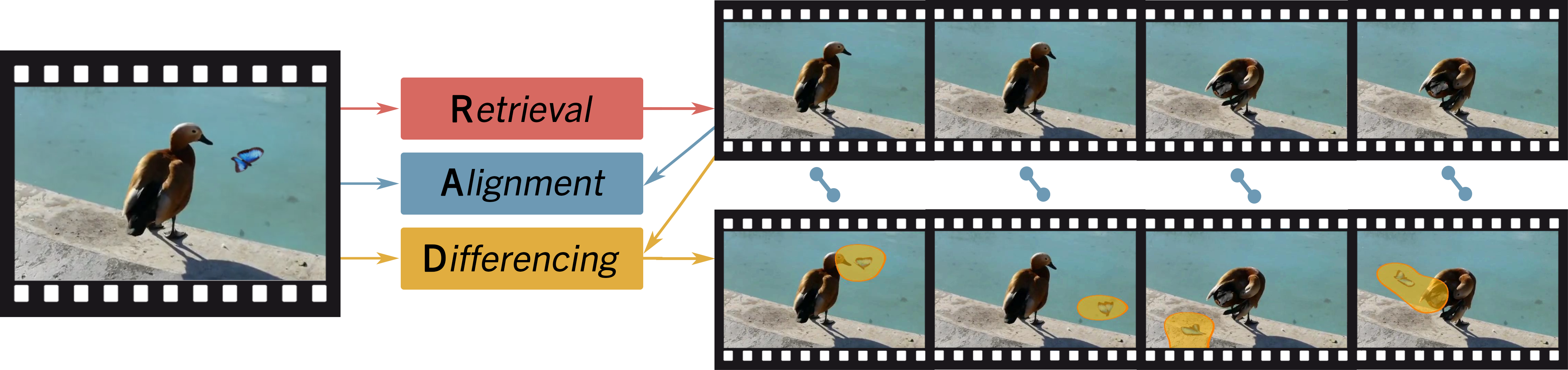}
\end{center}%
% \vspace{10pt}
}]

% \maketitle
\ificcvfinal\thispagestyle{empty}\fi

%%%%%%%%% ABSTRACT

\begin{abstract}

We propose VADER, a spatio-temporal matching, alignment, and change summarization method to help fight  misinformation spread via manipulated videos.  VADER matches and coarsely aligns partial video fragments to candidate videos using a robust visual descriptor and scalable search over adaptively chunked video content. A transformer-based alignment module then refines the temporal localization of the query fragment within the matched video.  A space-time comparator module identifies regions of manipulation between aligned content, invariant to any changes due to any residual temporal misalignments or artifacts arising from non-editorial changes of the content.  Robustly matching video to a trusted source enables conclusions to be drawn on video provenance, enabling informed trust decisions on content encountered. 
% We contribute a dataset for both temporal and spatial localization of video manipulations.

\end{abstract}

% \begin{figure*}
%     \centering
%     \includegraphics[width=1.0\linewidth]{figs/teaser.png}
%     \caption{teaser.}
%     % \vspace{-1em}
%     \label{fig:teaser}
% \end{figure*}

% Contribution summary for Scalable Fingerprinting:
% - Adaptive chunking technique to reduce size of the index and improve search time by leveraging temporal reduandancy in videos
% - Heuristic to aggregate the search results from multiple query chunks

%%%%%%%%% BODY TEXT
\section{Introduction}
\label{sec:intro}

Fake news and misinformation are major societal problems, much of it caused by manipulated videos shared to misrepresent events.  Detecting manipulation presents only a partial solution; most edited content is not misinformation \cite{ticks}.  Many emerging solutions, therefore, focus on {\em provenance} - determining the origins of content, what was changed, and by whom \cite{nguyen2021,ICN}.  Securely embedding such an audit trail inside video metadata can help consumers make better trust decisions on content, and is the basis of recent open standards (\eg C2PA) \cite{c2pa,origin}.  However, such metadata can be trivially stripped or replaced.

We propose VADER; a robust technique matching video fragments to a trusted database of original videos (such as those maintained by journalism or publishing organizations, per C2PA).  Once matched and temporally aligned, differences between the query-result pair are visualized to highlight any manipulation of the video (ignoring artifacts due to benign, \ie, non-editorial transformation due to renditions).  Any provenance metadata associated with the original may also be displayed.  Our technical contributions are:

\noindent \textbf{1. Scalable video fragment (R)etrieval.} We learn a robust self-supervised video clip descriptor invariant to various transformations of visual content. Video content is adaptively chunked and represented via these descriptors, which are aggregated and ranked to match and coarsely localize short video fragments within a large corpus of videos.
    
\noindent \textbf{2. Fine-grained temporal query (A)lignment.} We propose a Transformer based alignment method that performs fine-grained (\ie, frame level) localization of the matched video with the query fragment.  

\noindent  \textbf{3. Space-time (D)ifferencing to visualize manipulation.}  We learn a spatio-temporal model that `compares' the aligned query and matched video in order to identify regions of manipulation. The resulting heatmap ignores any discrepancies introduced due to residual frame misalignments or visual artifacts introduced by non-editorial changes such as quality, resolution, or format changes.

\noindent  \textbf{4. Video manipulation dataset.}  We contribute a dataset of 1042 professionally mANipulated videos and mAsK annotatIoNs (ANAKIN) to train and evaluate VADER. 
    
We show VADER's modular design meets three desirable properties we propose for practical video attribution: scalability in search; fine-grained localization of the query in the returned clip; and visualization of areas of manipulation.  We evaluate all three properties on our new dataset of manipulated video content (ANAKIN) in addition to standard video benchmarks (VCDB \cite{jiang2014vcdb} and Kinetics-600 \cite{kinetics}). We believe VADER is a promising step toward addressing fake news by helping users make more informed trust decisions on the video content they view online.%  complementing emerging open metadata standards for content provenance (\eg, C2PA \cite{c2pa}).    

%------------------------------------------------------------------------
\section{Related Work}
\label{sec:relwork}

Studies into video content authenticity generally fall into two camps: detection and attribution. 
Detection methods primarily evolve from the digital forensics perspective in detecting synthetic (or `deep fake') content or the presence of artifacts resulting from digital manipulation. 
The Deep Fake Detection Challenge (DFDC) \cite{kaggledf} catalyzed many works  \cite{mazaheri2022detection,zhao2021multi,zi2020wilddeepfake,guarnera2020deepfake} exploring the detection of facial manipulation in videos. 
Departures from natural image statistics could be identified via noise level analysis in the pixel domain \cite{ela} or via neural networks in the frequency domain \cite{dct2020}. These artifacts can be used to localize tampered regions \cite{zhang2019} or even determine which GAN architecture synthesized an image \cite{repmix}. 
Alternatively, content authenticity can also be studied as an anomaly detection problem, exploiting limitations in synthesis models such as lack of blinking in GAN-based deepfake videos \cite{blink} or local anomalous features introduced during the forgery process \cite{mantranet2019cvpr}.
However, not all misinformation is created via manipulation; imagery may be misattributed to tell a false story.

Attribution approaches, therefore, employ provenance-tracing methods to determine the origins of content and how it has been transformed from the original. 
Attribution is the focus of cross-industry coalitions (\eg, CAI \cite{cai}, Origin \cite{origin}) and emerging standards (\eg, C2PA \cite{c2pa}). 
A robust identifier may be embedded via metadata, watermarking \cite{devi2009,baba2009}, steganography \cite{yu2021responsible,weng2019high}) or obtained via perceptual hash of content (fingerprinting \cite{nguyen2021,Zhang2020manip,Bharati2021}).  
The identifier serves as a key to lookup provenance information in a database operated by a trusted source \cite{origin} or in a decentralized trusted database such as a blockchain \cite{archangel,bui2020archangel}.  
Challenges of fingerprinting include sensitivity to subtle manipulations and robustness to various perturbation sources, such as noise, compression, padding, enhancement, etc., during the re-sharing of media. These challenges can be addressed via object-centric tamper-sensitive hashing \cite{nguyen2021}, or multi-stage approaches \cite{ICN,zhang2020} that learn an identifier robust to both benign and editorial changes, opting instead to detect or highlight manipulation regions once matched. 
In all these works, provenance information can be exposed when a match is found via a pairwise comparison between the query and reference content, giving an advantage over blind detection methods. 

Content attribution for video is a largely unexplored topic due to challenges in fingerprinting arbitrary-length videos; the presence of noise and possible manipulations along the temporal dimension; video alignment and re-localization; and lacking real-world provenance datasets. 
Often, the query video is temporally cropped, and the video attribution task requires retrieving the original video and localizing the query within the matched candidate. 
A closely related problem is near-duplicate retrieval and video copy detection \cite{baraldi2018lamv,jiang2016partial,Douze2015CirculantTE,wu2007practical}.  
Early approaches perform fingerprinting with handcrafted features \cite{jiang2016partial,wu2007practical,douze2010image}, often omitting temporal information and resulting in false positives. 
In the era of deep learning, long video sequences have been modeled via recurrent networks for re-localization \cite{feng2018video}, hierarchical attention for attribution \cite{bui2020archangel} or metric learning for moment retrieval and partial copy detection \cite{baraldi2018lamv,jiang2016partial}. 
However, these works assume that the reference video is already identified. 
Several methods work on pairwise comparisons at inference time \cite{visil,han2021video} for fine-grained retrieval or alignment at the cost of scalability. VPN \cite{vpn} performs coarse-level video retrieval via a simple chunking algorithm followed by differencing with the best matched candidate, but does not take temporal cue into account.  
To our best knowledge, the most closely related state-of-art are  \cite{tan2022fast} for near-duplicate video search and \cite{baraldi2018lamv} for temporal alignment, both are later shown to under-perform our approach. 
Uniquely, our method supports video search, fine-grained alignment, and spatio-temporal manipulation visualization. 
Additionally, we contribute ANAKIN - a dataset of {\em professionally edited} videos with rich annotations, unlike existing near-duplicate/copy detection video benchmarks containing only simulated (\eg, TRECVID2008 \cite{trecvid2008}) or simple real-world transformations (\eg, VCDB \cite{jiang2014vcdb}).

% \begin{figure*}[t!]
%     \centering
%     \includegraphics[width=1.0\linewidth,height=6cm]{figs/full_arch.png}
%     \caption{}
%     \vspace{-1em}
%     \label{fig:arch}
% \end{figure*}

\section{Methodology}
\label{sec:method}
% \textcolor{red}{TODO (Alex): Methodology intro. Short description of all three components and how they form a pipeline}

Our proposed method consists of three main stages: retrieval, alignment, and differencing. Section~\ref{sec:method:retrieval} describes our video retrieval approach. We train an encoder in a self-supervised manner to extract robust video descriptors and use an adaptive chunking strategy to construct an index in a scalable manner.
We also introduce an approach for aggregating the retrieval results from multiple temporal chunks into a score for re-ranking, verification, and coarse temporal localization for downstream alignment. 
% A Priority Count Aggregation algorithm is used during query time to further improve retrieval results. 
Section~\ref{sec:method:algnment} describes the video alignment module. We propose a sequence matching transformer architecture, which is used to improve the alignment between the query and the retrieved video from chunk-level to a precise frame-to-frame matching. After alignment, the videos undergo pairwise comparison, described in Section~\ref{sec:method:detection}. Using a 3D CNN architecture, we produce a per-frame prediction of the presence of manipulations in query video via a heatmap.  

\subsection{Scalable Video Retrieval} \label{sec:method:retrieval}
\begin{figure}[t!]
    \centering
    \includegraphics[width=1.0\linewidth]{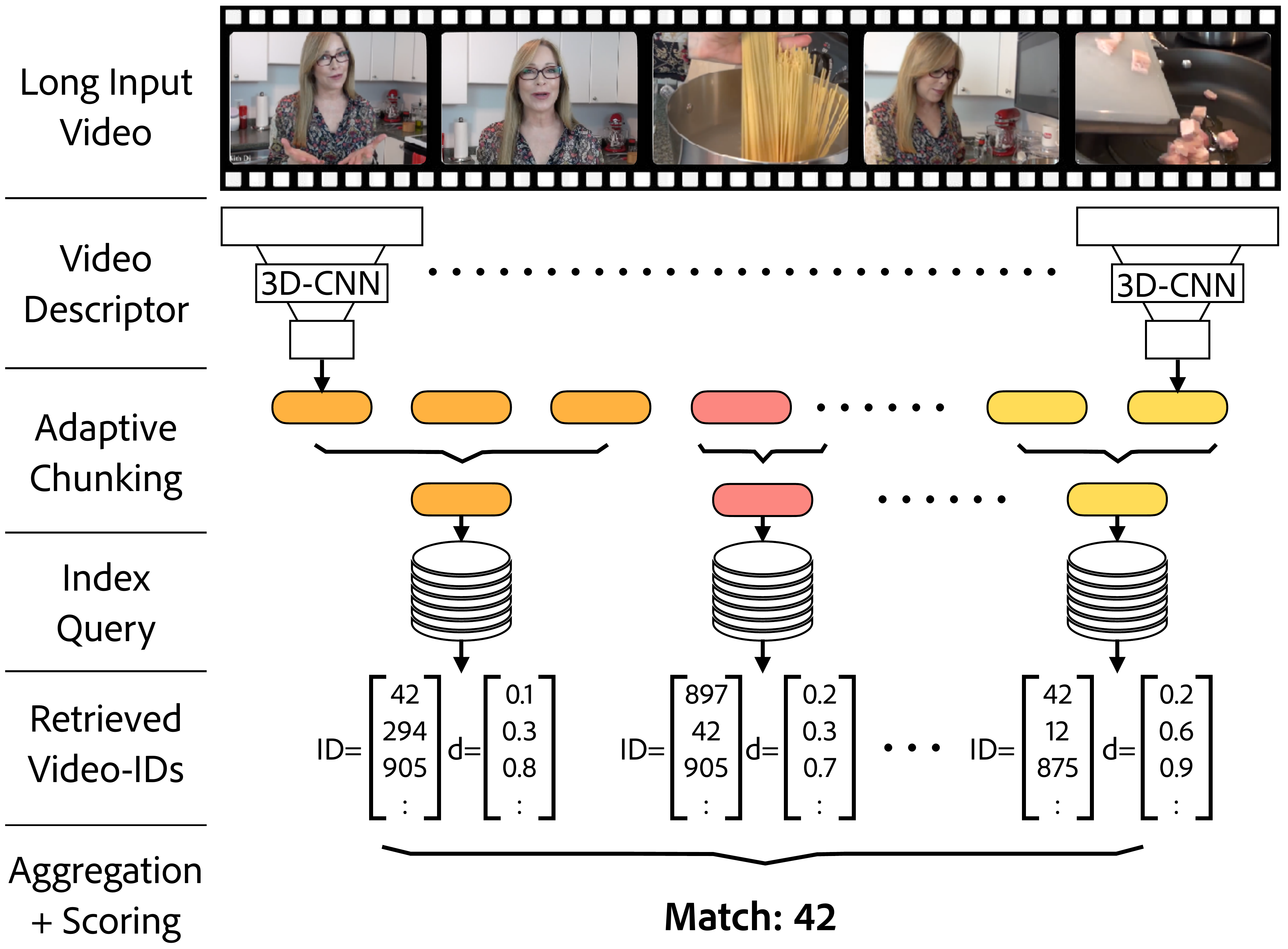}
    \caption{\textbf{Robust Scalable Retrieval of Long Videos.} 
     Given a long video, we extract robust descriptors for multiple non-overlapping temporal chunks. We exploit temporal redundancy via our adaptive chunking approach to reduce the number of descriptors required to represent a long video. Finally, we propose an approach to aggregate all the chunk-level retrieval results into a per-video score for re-ranking and verification. 
    }
    \label{fig:vfp}
    \squeezeup
\end{figure}

We aim to robustly match a video fragment to its corresponding entry in a large database, even when the query fragment only consists of a short segment (partial temporal matching) and undergoes various benign degradations and manipulations.
To address these challenges and to enable efficient retrieval and storage of long videos, our method (Figure~\ref{fig:vfp}) consists of (i) a robust self-supervised video clip descriptor, (ii) an adaptive chunking strategy to reduce the number of descriptors required to represent long videos, and (iii) aggregating chunk-level retrieval results into a single score for ranking and verification. 

\noindent \textbf{Robust Video Descriptor.}
Given an arbitrary-length video $\mathbf{v}$, we split into a non-overlapping sequence of chunks $v_i \in \mathbb{R}^{T \times H \times W \times 3}$. We empirically sample once every 4th frame and set $T=16$. 
Thus, each chunk represents about 3 seconds of video at 24fps. We train a 3D-CNN model \cite{Tran2017}, $F$, to encode each chunk to a compact descriptor. Concretely, $F(\mathbf{v})=[\nu_1, \ldots, \nu_n]$ where all $\nu_i$ are descriptors of fixed-sized chunks, $\nu_i = F(v_i) \in \mathbb{R}^{512}$. 
We require an $F$ robust to common visual degradations (benign, non-editorial changes)  and manipulations (c.f. subsec.~\ref{sec:aug}) and so employ three self-supervised training objectives as proposed in \cite{jenni2023audio}: 1) contrastive learning between augmented video clips, 2) contrastive learning between video and the corresponding audio track, and 3) temporal pretext tasks, \eg, the prediction of video playback speed and direction. 
The Kinetics-600 dataset \cite{kinetics} is used to pre-train $F$ with these self-supervised objectives.

\noindent \textbf{Adaptive Long-Form Video Indexing.}
Representing videos as a sequence of per-chunk features would lead to a linear scaling of the index size as a function of video duration.
However, as videos have high temporal redundancy, \ie, consecutive video chunks often have similar semantics, we propose an adaptive chunking strategy to reduce the number of descriptors for each video.
Our approach defines a threshold $\tau$ for consecutive chunk similarity above which we aggregate similar chunks. 
% We first extract a large number of chunk descriptors $F(v_i) \in \mathcal{V}$ from a set of diverse long-form videos.
% We then cluster these descriptors using $k$-Means and use the cluster assignments $\rho(\nu)$ of consecutive video chunks for adaptive chunking. 
% Adaptive chunking results in shorter video descriptors   $\hat{F}(\mathbf{v})=\left[\Bar{\nu}_{0:i_1}, \ldots, \Bar{\nu}_{i_m:n} \right]$, where aggregated descriptors are given by:
% \begin{equation}
%     \Bar{\nu}_{j:k}=\frac{1}{k-j} \sum_{i=j+1}^k \nu_i.
% \end{equation}
Adaptive chunking results in shorter video descriptors  
\begin{equation}\label{eq:ada_chunk}
    \hat{F}(\mathbf{v})=\left[\Bar{\nu}_{0:i_1}, \ldots, \Bar{\nu}_{i_m:n} \right],
\end{equation}
where aggregated descriptors are given by $\Bar{\nu}_{j:k}=\frac{1}{k-j} \sum_{i=j+1}^k \nu_i.$
The chunk boundaries $i_j$ in Eq.~\ref{eq:ada_chunk} are given by all the indices $i \in \{1, \ldots, n-1\}$ where $d(\nu_i, \nu_{i+1})<\tau$, \ie, where the cosine similarity between consecutive chunks is below the chosen threshold $\tau$.
In practice, we chose $\tau$ by computing consecutive chunk similarities for a large and representative set of long videos and set $\tau$ to be the $k$-th percentile of the resulting distribution. 
Note that a lower $\tau$ corresponds to more aggressive chunking and that the chosen percentile roughly corresponds to the compression rate.
We also explore the use of clustering to define the chunk boundaries, pre-computing a fixed clustering of the feature space via k-means and defining chunk boundaries as the indices where the cluster assignments change, \ie,  $\rho(\nu_i)\neq \rho(\nu_{i+1})$ for cluster assignment $\rho$.
% Note that the number of clusters $k$ controls the strength of the effect, \ie, fewer clusters will lead to a stronger compression through adaptive chunking. 

\noindent \textbf{Aggregating and Ranking Search Results.} With each video being represented by a sequence of adaptively chunked descriptors (Eq.~\ref{eq:ada_chunk}), a question arises of how to store, query, and rank results for videos containing multiple temporal chunks. 
We store each descriptor $\Bar{\nu}$ along with its video ID and temporal position in an efficient inverted file index with Product Quantization (IVFPQ) \cite{faiss}. 
Given an adaptively chunked sequence of query descriptors at query time, we propose a method to aggregate multiple chunk-level retrieval results to rank the retrieved videos according to how likely they are to contain the query fragment. 
% We call our approach Priority Count Aggregation, which is illustrated in Algorithm~\ref{alg:1}. 
% To summarize, the procedure iterates over retrieval ranks, sorts video IDs in each rank by count (breaking ties using minimum chunk distance), and aggregates video IDs in a ranked list.
To this end, let $\mathbf{S}\in \mathbb{N}^{m \times r}$ and $\mathbf{D}\in \mathbb{R}^{m \times r}$ represent the chunk-level retrieval results for a query $F(\mathbf{v})\in \mathbb{R}^{m \times 512}$, \ie, $S_{ij}$ is the video-ID for chunk $i$ at rank $j$ and $D_{ij}$ the cosine similarity between the query and retrieved chunk descriptor.
Let us further define $\operatorname{rank}_i(s) = \min \{ j | S_{ij}=s \}$ as the rank of the first chunk belonging to video $s$ for query chunk $i$, and $\sigma_i=\operatorname{std}(\{ D_{i1}, \ldots,  D_{ir}\} )$ as the standard deviation of the chunk similarities for query chunk $i$.
We then assign a weight to each combination of query chunk $\Bar{\nu}_i$ and video-ID $s$ through 
\begin{equation}\label{eq:chunkweight}
    % \operatorname{score}(s) = \sum_{i=1}^{m} \sum_{j=1}^{r} \mathbbm{1}\{ \}
    % \operatorname{score}_i(s) = \sum_{i=1}^{m}  \mathbbm{1}\{\}  \frac{\sigma_i}{\operatorname{rank}_i(s)^\lambda}
    w_i(s) = \begin{cases}
  \frac{\sigma_i}{\operatorname{rank}_i(s)^\lambda}  & \text{if } s \in \{S_{i1}, \ldots, S_{ir} \} \\
  0 & \text{otherwise} \\
  \end{cases}.
\end{equation}
Per Eq.~\ref{eq:chunkweight}, we assign higher weights when video $s$ appears at a low rank and when the chunk similarities have higher standard deviation (indicating more distinctive chunk descriptors). 
The parameter $\lambda$ controls the relative importance of higher rank retrievals and is set to $\lambda=2$ by default.
Finally, the retrieved videos are ranked according to the score
\begin{equation}\label{eq:vidscore1}
    % \operatorname{score}(s) = \frac{1}{m}  \sum_{i=1}^{m} w_i(s).
    \operatorname{score}(s) =  \frac{1}{m}  \sum_{i=1}^{m} w_i(s) \cdot d_i(s),
\end{equation}
where $d_i(s)$ is the cosine similarity between query chunk $i$ and the most similar retrieved chunk from video $s$.

\noindent \textbf{Coarse Temporal Alignment.}
Next, we describe how to coarsely localize a query segment in a retrieved video and reduce the search space for downstream temporal alignment. 
Let $t_i(s)$ represent the starting time of the first retrieved chunk of video $s$ for query $i$. 
To coarsely localize the query in the retrieved videos, we propose to use the weighted sum of the retrieved chunk starting times
\begin{equation}\label{eq:vidscore2}
    \operatorname{start}(s) =  \sum_{i=1}^{m} \hat{w}_i(s) \big( t_i(s) - \Delta t_i \big) ,
\end{equation}
where $\Delta t_i$ is the time between the first and $i$-th query chunk, and the weights $\hat{w}_i(s)$ are per-video normalized versions of the weights in Eq.~\ref{eq:chunkweight}, \ie, $\hat{w}_i(s) = \frac{w_i(s)}{\sum_{j=1}^m w_j(s)}$.
% \begin{equation}
% \hat{w}_i(s) = \frac{w_i(s)}{\sum_{j=1}^m w_j(s)}.
% \end{equation}
\\

% \begin{algorithm}[!t]
% \SetAlgoLined
% \small
% \KwIn{Matrix of per-chunk retrieved videos $\mathbf{s}\in \mathbb{N}^{m \times r}$, where $s_{ij}$ is the ID for chunk $i$ at rank $j$. 
% } 
% \KwOut{List of aggregated ranked video IDs $\mathbf{a}$}
% Initialize $\mathbf{a}=[]$ \;
% \For{$i=1\ldots r$}{
%         \# sortByCount \textit{breaks ties using min. distance}\;
%         $\mathbf{c} \leftarrow $ sortByCount($[s_{1i}, \ldots, s_{mi}]$)\;
%         $\mathbf{a} \leftarrow $ addMissing($\mathbf{a}$, $\mathbf{c}$)\;        
%         % \For{$s \in \mathbf{c}$}{
%         %      \If{$s \not\in \mathbf{a} $}{
%         %          $\mathbf{a} \leftarrow  \mathbf{a} + s$
%         %      }
%         % }
% }
% \caption{\small Priority Count Aggregation}
% \label{alg:1}
% \end{algorithm}

\subsection{Fine-Grained Video Alignment}\label{sec:method:algnment}

\begin{figure*}[t!]
    \centering
    \includegraphics[width=1.0\linewidth]{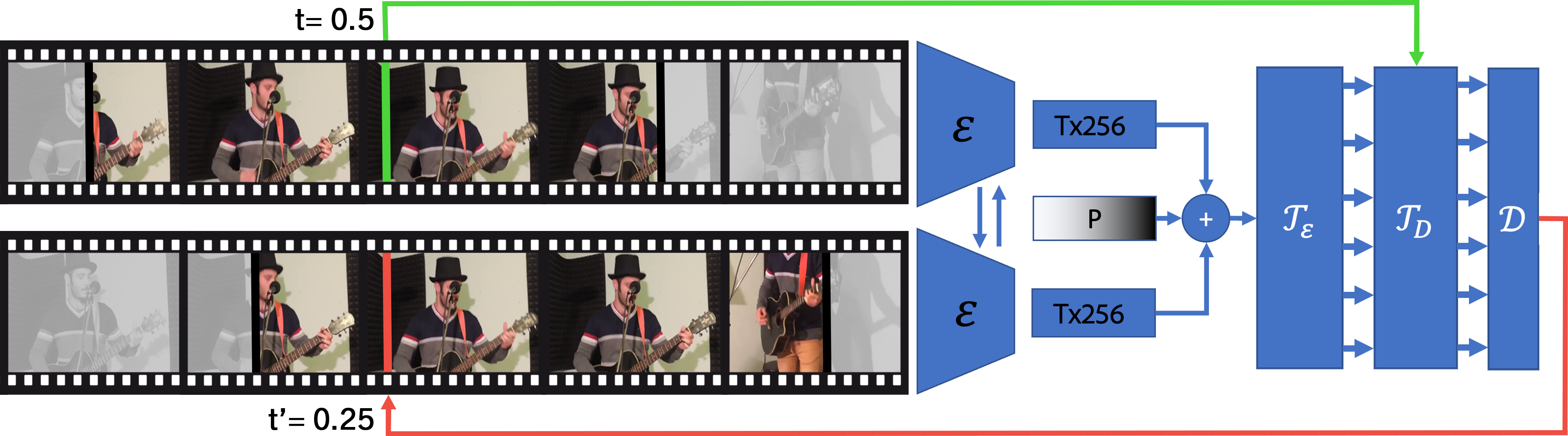}
    \caption{Architecture diagram of the proposed alignment transformer $\mathcal{T}$. Sequences of robust fingerprints are encoded using $\varepsilon(.)$\cite{ICN} from the query $V$ and candidate videos $V'$. Together with a positional encoding, they are fed into a transformer encoder $\mathcal{T_E}$. Given query times $t$, the decoder transformer $\mathcal{T_D}$ provides an output, which is interpreted by an MLP ${D}$ to match query times  with corresponding times $t'$.}
    \vspace{-1em}
    \label{fig:alignment_arch}
\end{figure*}

We perform fine-grained video alignment to find corresponding matches between frames of the query video fragment, and the frames of the video clip identified and coarsely aligned by the prior retrieval step. This finer-grained alignment is necessary for subsequent manipulation detection, described in Section~\ref{sec:method:detection}. We propose a deep neural network architecture for alignment robust to  manipulations and temporal augmentations (Figure \ref{fig:alignment_arch}).

\noindent \textbf{Model Architecture.}
Inspired from \cite{COTR, DETR, vit2020}, we implement a transformer network $\mathcal{T}$ that, given a query video $V$, a retrieved video $V'$ and a normalized time coordinate $t \in [0,1]$ is tasked to output time $t' \in [0,1]$ such that frame $V(t)$ and $V'(t')$ are temporally aligned. Unlike \cite{COTR, DETR, vit2020}, which work on images, we uniquely leverage a transformer for the pair-wise video alignment task.

For each video frame, we extract a robust $256$ dimensional feature embedding using a frame encoder $\varepsilon$. A video consisting of $T$ frames can be represented as a $T \times 256$ sequence of frame descriptors $\varepsilon(V) = [\varepsilon(V_1), \varepsilon(V_2), \ldots, \varepsilon(V_T)]$. Since our approach is agnostic of the feature choice, we follow \cite{baraldi2018lamv, Douze2015CirculantTE, TMK} and use RMAC \cite{RMAC} features and explore other options in the supmat. 

We subsample to the length of $20$ and concatenate together the feature sequences of two videos $V$ and $V'$. A vector of positional encodings $P$ is added to the $40 \times 256$ sequence before it is used as an input for the transformer encoder $\mathcal{T_E}$. The output of the transformer encoder, together with the positional encoding of the query time $\mathcal{P}(t)$, is passed into the transformer decoder $\mathcal{T_D}$. Then, a Multi-Layer Perceptron (MLP) $D$ is used to interpret the output of the transformer decoder and obtain a prediction of the corresponding time $\hat{t}'$.
To summarize, inputs to the model are $c = [\varepsilon(V), \varepsilon(V')] + P$ and predictions computed as
\begin{equation}
  \hat{t}' = \mathcal{T}(t | V, V') = D\Big(\mathcal{T_D}\big(\mathcal{P}(t), \mathcal{T_E}(c)\big)\Big).
  \label{eq:alignment1}
\end{equation}

As in \cite{COTR}, we use a linear increase in frequency for the positional encoding instead of log-linear to yield a more stable optimization. The positional encoding for a given time $t$ is defined as $\mathcal{P}(t) = [\sin(1\pi t), \cos(1\pi t), \ldots, \sin(k\pi t), \cos(k\pi t)]$, where $k=128$ results in a sequence of length $256$.

\noindent \textbf{Learning Objectives.}
The network is trained to minimize two losses: mean squared error (MSE) $\mathcal{L_{\mathit{MSE}}}$ and a cycle loss $\mathcal{L_{\mathit{cycle}}}$. The first loss is defined as the error between the ground truth time $t'$ and the network's prediction $\hat{t}'$ as: $\mathcal{L_{\mathit{MSE}}} = \lVert t' - \hat{t}' \rVert^2_2.$ The cycle loss ensures that the network is able to reconstruct the query time $t$ using its own prediction of the corresponding time $\hat{t}'$:  $\mathcal{L_{\mathit{cycle}}} = \lVert t - \mathcal{T}(\hat{t}' | V', V) \rVert^2_2.$ The total loss is $\mathcal{L} = w_m \mathcal{L_{\mathit{MSE}}} + w_c \mathcal{L_{\mathit{cycle}}}$ where loss weights $w_m=0.6, w_c=0.4$ are set empirically.

\noindent \textbf{Frame Sampling.}
During training, we subsample 20 frames from each video. From the non-manipulated video $V$ of length $N$ we select a set of frames $[V_{s + 0k}, V_{s + 1k} \ldots V_{s + 19k}]$ where the starting frame $s$ and stride $k$ can be any combination of integers that satisfies the condition $0 \leq s < s + 19k \leq N$, chosen at random. For the second video, the starting frame is shifted by $19k * \alpha$ rounded to the nearest integer, where $\alpha \in [0, 0.25]$. This ensures that both sequences' coverage of the video overlaps by at least 50\%. 
Additionally, the indices of the second video are subjected to random perturbations $\mathbf{g}_i \sim \mathcal{N}(\mu,\,\sigma^{2})$ rounded to the nearest integer (we set $\mu=0$ and $\sigma=4$): $[V'_{s' + 0k + \mathbf{g}_0}, V_{s' + 1k + \mathbf{g}_1} \ldots V'_{s' + 19k + \mathbf{g}_{19}}]$.
The set of pairs of query times $t$ and corresponding times $t'$ consists of normalized time coordinates of $V$ that overlap with $V'$ and thus can vary in size between $10$ and $20$ pairs.
During inference, to ensure no loss in precision, no subsampling is performed but rather each consecutive set of 20 frames is evaluated in a sliding  window fashion.

\noindent \textbf{Implementation Details.}
The transformer encoder $\mathcal{T_E}$ is implemented as 6 layers of 8-head self-attention modules, and the transformer decoder $\mathcal{T_D}$ as 6 layers of 8-head encoder-decoder attention modules. There is no self-attention in the decoder to make queries independent of each other.
The MLP module $D$ consists of $d_{mlp}=6$ fully-connected layers with a hidden dimension size of $h_{mlp}=2048$. These values were determined using grid-search with $d_{mlp} \in [1, 10]$ and $h_{mlp} \in [2^5, 2^{12}]$.

\subsection{Video Manipulation Detection}
\label{sec:method:detection}

We utilize the convolution network inflation method \cite{I3D} to extend the image comparator network (ICN) \cite{ICN} to the video domain. We later show that encoding multiple frames via a 3D-CNN architecture provides more temporally stable differencing visualizations than single-frame methods like ICN. Additionally, introducing slight misalignments between the frames during training results in a model resilient to alignment errors that may occur during the video alignment step.

\begin{figure}[t!]
    \centering
    \includegraphics[width=0.87\linewidth]{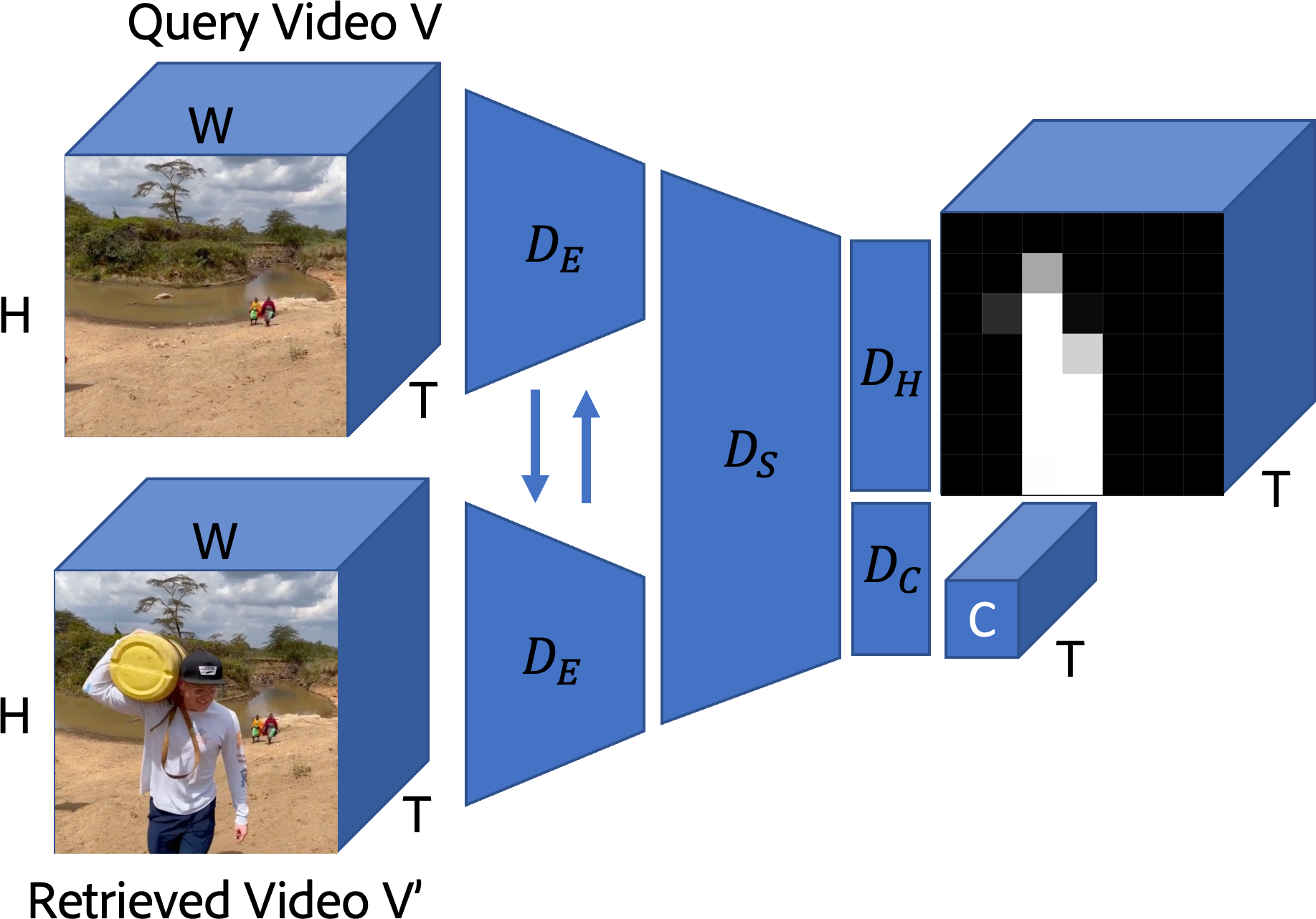}
    \caption{ Architecture of the VADER video differencing module. Video tensors are encoded separately with $\mathcal{D}_E$, then a joint embedding is learned in $\mathcal{D}_S$. The embedding is used to produce a heatmap, localizing frame manipulations. A binary classification is also made per frame, marking it as manipulated or not. For non-manipulated frames, the heatmap is zeroed out.}
    \vspace{-1em}
    \label{fig:breakdown}
\end{figure}

\noindent \textbf{Model Architecture.} Our model (Figure \ref{fig:breakdown}) takes as input two video clips of size $T\times C \times W \times H = 16\times 3 \times 224 \times 224$. The model output is a $16\times 7 \times 7$ grid which, when interpolated to the original frame size, serves as a heatmap that highlights the edited region of the video. We inflate ICN using the approach described in \cite{I3D}, where we substitute 2D convolutions with weights repeated along the time dimension to produce 3D convolutions. In our case, the stride along the time dimension is always kept at $1$, resulting in an output with the same temporal resolution as the input - one heatmap per frame.

We train a video differencing 3D-CNN $\mathcal{D}$ that, given a query video $V'$ and a retrieved and aligned video $V$, outputs a heatmap $h \in \mathbb{R}^{16 \times 7 \times 7}$ and a binary classification score $c$, signifying whether the query video has been manipulated. First, both videos are passed separately through a shared feature extractor $\mathcal{D}_E$:
\begin{align}
    z_0=\mathcal{D}_E(V);\; z_0'=\mathcal{D}_E(V') \in \mathbb{R}^{T \times H'\times W' \times C}.
\end{align}
% where $H'$, $W'$ and $C$ are the new height, width and feature dimension respectively. Note that the temporal dimension $T$ remains unchanged.
$\mathcal{D}_E$ contains three 3D convolution layers separated by ReLU, batch norm, and max pooling. It  outputs features at $\frac{1}{4}$ resolution ($H'=H/4, W'=W/4$ and we set $C=64$). Features are concatenated to form an input of size $T\times H' \times W' \times 2C$ to the next module $\mathcal{D}_S$ producing $z=\mathcal{D}_S([z_0, z_0']) \in \mathbb{R}^{T \times 256}$. $\mathcal{D}_S$ contains 4 residual blocks \cite{he2016deep} then an FC layer.

\noindent \textbf{Learning Objective.} To correctly recognize whether the query video was manipulated and produce accurate localization masks of the manipulated regions, we train the differencing module with two losses. First, the similarity loss $\mathcal{L}_S$ minimizes the cosine distance between the manipulation heatmaps derived from $z$ and the ground truth heatmaps. We produce the heatmap at resolution $T \times n\times n$ from $z$ via a FC layer $\mathcal{D}_H(z) \in \mathbb{R}^{T\times n^2}$ and compute the loss:
\begin{align}
    \mathcal{L}_S = \frac{1}{T} \sum_{t=1}^{T}{1 - \frac{\mathcal{D}_H(z_t)\cdot G_t}{\left|\mathcal{D}_H(z_t)\right|\left|G_t\right|})}
\end{align}
where $G_t$ is the ground truth manipulation heatmap of the $t$-th frame. We define the output heatmap resolution $n=7$ during training. At test time, the $7\times 7$ heatmaps are interpolated using bicubic interpolation to the original resolution $H\times W$ and super-imposed on the query video. The heatmaps are continuous but can be thresholded for more intuitive visualization. Additionally, we apply a classification loss $\mathcal{L}_C$ computed as binary cross-entropy loss:
\begin{align}
    \mathcal{L}_C &= \frac{1}{T} \sum_{t=1}^{T}{-\log {\frac{e^{c_{t,y_t}}}{\sum_{i=1}^2{e^{c_{t,i}}}}}}
\end{align}
where $c = \mathcal{D}_C(z) \in \mathbb{R}^{T \times 2}$ with a FC layer $\mathcal{D}_C$, and $y_t$ is the classification target, indicating if the $t$-th frame of the query contains manipulations.

The total loss is $\mathcal{L}_D = w_s \mathcal{L}_S + w_c \mathcal{L}_C$ where loss weights $w_s=w_c=0.5$ are set empirically.

\section{Experiments and Discussion}

We evaluate the performance of the VADER retrieval, alignment and detection stages.  We use three datasets: ANAKIN, Kinetics-600, and VCDB (Sec~\ref{sec:datasets}).  ANAKIN is our end-to-end dataset used to evaluate all three stages.  Kinetics is additionally combined with ANAKIN to evaluate retrieval scalability.  VCDB is a benchmark additionally used to baseline the accuracy of our retrieval method.

\subsection{Datasets} \label{sec:datasets}

\begin{figure*}[t!]
    \centering
    \includegraphics[width=0.97\linewidth,height=5.5cm]{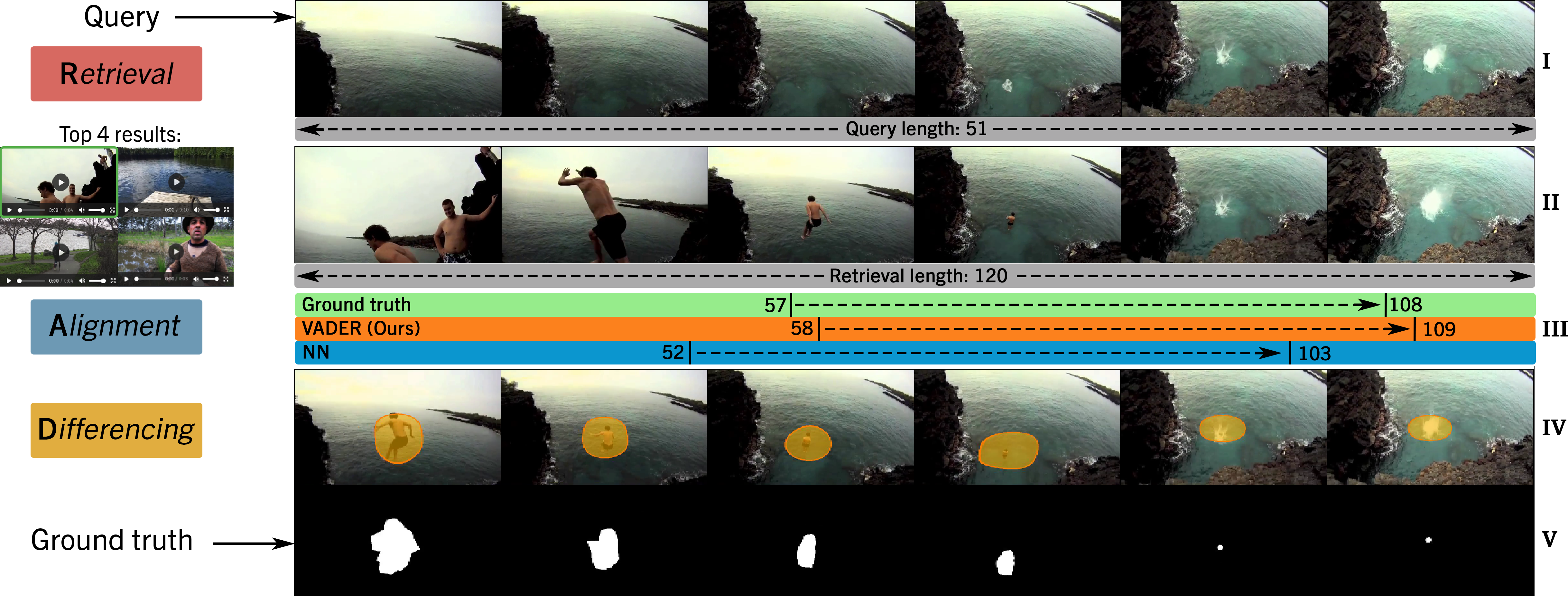}
    \caption{An example of video manipulation detection using VADER. \rom{1}: a video consisting of 51 frames is used as a query. \rom{2} : a correct original video with 120 frames is retrieved (top-4 results inset - left). \rom{3}: alignment module is used to temporally localize the query withing the retrieved video, fifth 1 frame error. \rom{4}: differencing module highlights the manipulated region (superimposed here on original video for clarity). \rom{5}: ground truth mask of the manipulated region from ANAKIN.}
    \label{fig:example}
\end{figure*}

\noindent \textbf{ANAKIN\footnote{This dataset will be released as CC-BY 4.0 upon publication.}} is a new dataset of mANipulated videos and mAsK annotatIoNs, which we introduce. To our best knowledge, ANAKIN is the first real-world dataset of professionally edited video clips paired with source videos, edit descriptions, and binary mask annotations of the edited regions. ANAKIN consists of 1042 videos in total, including 352 edited videos from the VideoSham \cite{sham} dataset, plus 690 new videos collected from the Vimeo platform. 
For each video, professional video editors were given a task (\eg, ``change the color of the shirt of the person running in the background") and a time interval on which to perform the task. Editorial tasks include: frame-level manipulations (duplication, reversing, dropping, speeding up, slowing down), audio changes (addition, replacement), splicing (adding objects, text, or entities), inpainting (removing objects, text or entities), and swap (background or color). The mean length of the original videos is 120 seconds, and the mean length of the edited clips is 5s. The video editors were also asked to provide binary segmentation masks as the ground truth for their edited regions. Examples of original and manipulated videos along with their segmentation masks, can be found in Figure~\ref{fig:example}. 

\begin{table}[t!]
\centering
\begin{adjustbox}{max width=\columnwidth}
\begin{tabular}{@{}l@{\hspace{1em}}c@{\hspace{1em}}c@{\hspace{1em}}c@{\hspace{1em}}c@{\hspace{1em}}c}
\toprule
   &\textbf{Comp.}  &  \multicolumn{2}{c}{\textbf{Aug.-K600}}  &  \multicolumn{2}{c}{\textbf{ANAKIN}} \\
\textbf{Chunking} & \textbf{Rate} & \textbf{R@1} & \textbf{R@5}  & \textbf{R@1} & \textbf{R@5}  \\ \midrule
All &  $32\times$  &   67.7  &  71.2 & 81.9  & 84.6 \\
None &   -  &  \textbf{76.6}  & 81.3 & 96.9  & 98.6 \\
\midrule
Constant &  $2\times$ & 75.0  &  80.0  & 96.5 & 97.9 \\
Constant &  $4\times$ & 72.6  &  77.8  & 94.7  & 95.9 \\
\midrule
Adaptive (k-means) &   $2\times$  & 76.0  & \textbf{81.9} &  96.9 & 98.2  \\ 
Adaptive (sim-thresh) &   $2\times$  & \textbf{76.6}  & 81.5 & \textbf{97.2} & \textbf{98.8} \\ 
Adaptive (sim-thresh) &   $4\times$  & 74.0 & 79.8  & 96.1 & 95.9  \\ 
\bottomrule 
\end{tabular}
\end{adjustbox}
\caption{
\textbf{Adaptive Chunking.}
We compare the retrieval performance of our proposed adaptive chunking to baselines retaining all chunks (None), aggregating all the chunks per video (All), and constant chunking (averaging $k\in\{2, 4\}$ consecutive chunks).
}\label{tab:adachunk}

\end{table}

\begin{table}[t!]
\centering
\begin{adjustbox}{max width=\columnwidth}
\begin{tabular}{@{}l@{\hspace{1em}}c@{\hspace{1em}}c@{\hspace{1em}}c@{\hspace{1em}}c@{\hspace{1em}}c@{\hspace{1em}}c@{\hspace{1em}}c}
\toprule
    &  \multicolumn{3}{c}{\textbf{Aug.-K600}}  &  \multicolumn{3}{c}{\textbf{ANAKIN}} \\
\textbf{Scoring}  & \textbf{R@1} & \textbf{R@5} & $\mathbf{F_1}$  & \textbf{R@1} & \textbf{R@5}  & $\mathbf{F_1}$ \\ \midrule
Count &    62.9  &  77.2  & 52.0  &  88.6 & 98.3 &  51.6 \\
Max. Sim. &   74.8   &  79.0 &  50.5  & \textbf{97.0} & 98.5 & 93.5 \\
Uniform &   75.2   &  79.6 & 64.3 & 96.9 &  98.5 & 94.6  \\
\midrule
Ours (w/o $\sigma$)   &   \textbf{76.6}   &  81.2  &  84.2   &  96.9 &  \textbf{98.6}  & 97.6 \\
Ours   &   \textbf{76.6}   &  \textbf{81.7}  &  \textbf{88.0} &  96.9 &  \textbf{98.6}  & \textbf{97.7} \\
\bottomrule 
\end{tabular}
\end{adjustbox}
\caption{
\textbf{Retrieval Aggregation and Scoring.}
We compare our scoring function (Eq.~\ref{eq:vidscore1}) to several alternative formulations. }\label{tab:scoring}
% }
\end{table}

\begin{table}[t!]
\centering
% \resizebox{\textwidth}{!}{%
\begin{tabular}{@{}l@{\hspace{1em}}c@{\hspace{1em}}c@{\hspace{1em}}c}
\toprule
\textbf{Method} & \textbf{SP} & \textbf{SR} & \textbf{$\mathbf{F_1}$-Score}  \\ \midrule
CNN \cite{jiang2016partial}  & -  & -  & 0.6503 \\
TH+CC+ORB \cite{guzman2016towards}  & 0.5052  &  0.9294 & 0.6546  \\
LAMV \cite{baraldi2018lamv}  & -  & -  & 0.6870  \\
Q-Learning \cite{guzman2019partial} &  0.8829 &  0.7355 & 0.8025 \\
BTA \cite{zhang2016effective} &  0.7600  &  0.7500 &  0.7549 \\
VSAL \cite{han2021video}  &   \textbf{0.8971} & 0.8462  & 0.8709 \\
FPVCD \cite{tan2022fast}   & -  & -  &   0.8764  \\
\midrule
Ours &  0.8652 & \textbf{0.9753} &    \textbf{0.9167} \\ 
\bottomrule 
\end{tabular}
\caption{
\textbf{Video Copy Detection on VCDB.}
We compare to prior results on VCDB reporting maximum segment-level $F_1$-scores and associated precision (SP) and recall (SR) as defined in \cite{jiang2014vcdb}.}\label{tab:vcdb}
% }
\squeezeup
\end{table}

ANAKIN is used to train and evaluate VADER's alignment and manipulation detection models in Sec.~\ref{sec:align} and \ref{sec:manip}, with an 80:20 train-test split in both tasks. Since we focus on visual manipulation detection, videos in which manipulation is difficult to visualize, \eg, audio replacement, are excluded from our manipulation detection experiments.   
% First, we use the edited clips as queries to evaluate video retrieval performance. Then, we assess the performance of the proposed video alignment approach by exploiting the known exact start and end times of the edited clips within the longer retrieved original videos. Finally, binary mask annotations are used to evaluate manipulation detection and localization performance.

% To pre-train our video manipulation detection model, we employ PSBattles \cite{psBattles}; a dataset of `in-the-wild' images manipulated by both amateurs and professional artists. The dataset was collected from the `Photoshopbattles’ forum on  Reddit by \cite{psBattles}, cleaned and annotated by \cite{ICN} with bounding boxes highlighting manipulation regions. The final dataset has $\sim$7000 original and 24000 photoshoped images which were used to train a single-frame ICN before being inflated to the 16-frame 3D ICN as described in sec.~\ref{sec:method:detection}.

\noindent \textbf{(Aug.-)K600} consists of 60,000 videos in the Kinetics-600 \cite{kinetics} test set.
We use the \emph{full-length} source videos (not just the 10-second labeled clips) to build the search index for our scalable retrieval experiment. This results in a search index containing approximately 4M descriptors (without adaptive chunking).  Retrieval experiments on ANAKIN also include these descriptors as distractors. We also generate a query set \textbf{Aug.-K600} of benign augmented query clips that are temporally truncated  to test partial retrieval.    In addition to ANAKIN, the Kinetics-600 training partition is used to train the robust descriptors for the retrieval stage.

\noindent \textbf{VCDB} \cite{jiang2014vcdb}
is a well-established video copy detection benchmark. It consists of over 9K pairs of labeled in-the-wild copied video fragments. 
We use the `core' dataset to make an additional evaluation of our retrieval method versus partial video copy detection methods. 

\noindent \textbf{Benign (Non-Editorial) Augmentations.} \label{sec:aug}
During the training of all stages of VADER, we apply random video augmentations to clips using AugLy \cite{augly} to simulate non-editorial (benign) transformation of content during online distribution. The same set of random augmentations is applied to query fragments, where indicated in the evaluation of each of the 3 VADER stages.  These augmentations encompass  in-place manipulations (color jittering, blurring, noise, etc.), geometric transforms (cropping, rotations, flipping), and temporal cropping (see supmat).

\subsection{Scalable Video Fragment Retrieval}
% \textcolor{red}{TODO: Show retrieval performance on the edited dataset. Compare adaptive vs. constant chunking. Compare ranking/aggregation heuristics.}

\noindent \textbf{Evaluation Metrics.}
To evaluate the retrieval performance, we report recall-at-$k$ (R@k) and the maximum fragment-level F1 score as defined in \cite{jiang2014vcdb}.
While recall indicates how well our method can rank the retrieved videos, the F1 score captures the method's ability for coarse temporal localization (detections must overlap with ground truth) and its ability to discriminate between true and false matches.

\noindent \textbf{Adaptive Chunking.}
We compared our adaptive chunking to several baselines regarding retrieval performance and achieved compression rate (\ie, the  average reduction in descriptor length) in Table~\ref{tab:adachunk}.
We consider the following baselines in our comparison:
\noindent \textbf{All} - In this case, each video is represented via a single aggregated descriptor, \ie, $\hat{F}(\mathbf{v})=[\Bar{\nu}_{0:n}]$. This naive single-descriptor approach corresponds to the maximal achievable compression via chunking.
\noindent \textbf{None} - No chunking is applied, and as a result, no additional compression is achieved. 
\noindent \textbf{Constant} - In these variants, the temporal chunking is not adaptive but instead aggregates a fixed number of $k\in \{2,4\}$ consecutive descriptors, \ie, $\hat{F}(\mathbf{v})=\left[\Bar{\nu}_{0:k}, \ldots, \Bar{\nu}_{n-k:n} \right]$.

\begin{table*}[t!]
\centering
% \resizebox{\textwidth}{!}{%
\begin{tabular}{lllllllllllll}
\hline
\multirow{2}{*}{\textbf{Method}} & \multicolumn{3}{c}{\textbf{clean}} & \multicolumn{3}{c}{\textbf{clean + benign}} & \multicolumn{3}{c}{\textbf{manip}} & \multicolumn{3}{c}{\textbf{manip + benign}} \\
 & \textbf{@0.1s} & \textbf{@1s} & \textbf{@10s} & \textbf{@0.1s} & \textbf{@1s} & \textbf{@10s} & \textbf{@0.1s} & \textbf{@1s} & \textbf{@10s} & \textbf{@0.1s} & \textbf{@1s} & \textbf{@10s} \\ \hline
VADER & \textbf{69.5} & \textbf{91.6} & \textbf{97.4} & \textbf{64.2} & \textbf{78.4} & \textbf{93.2} & \textbf{54.2} & \textbf{81.1} & \textbf{94.7} & \textbf{53.7} & \textbf{74.2} & \textbf{91.6} \\
LAMV \cite{baraldi2018lamv} & 54.7 & 75.3 & 87.9 & 44.7 & 68.9 & 85.3 & 38.4 & 66.3 & 86.3 & 30.0 & 59.5 & 83.2 \\
CTE \cite{Douze2015CirculantTE}& 4.8 & 25.4 & 67.7 & 4.2 & 20.1 & 67.2 & 4.8 & 21.2 & 68.3 & 3.7 & 19.6 & 68.3 \\
TMK \cite{TMK}  & 2.6 & 16.8 & 69.5 & 1.6 & 18.4 & 63.2 & 1.6 & 15.8 & 66.8 & 3.2 & 17.9 & 59.5 \\
\hline
\end{tabular}
% }
\caption{Alignment module evaluation on ANAKIN dataset. We report percentage of videos aligned better than a  threshold value (0.1s, 1s, 10s). VADER outperforms baseline methods in all four query sets and shows more resilience to benign transformations.}
\label{tab:align}
\end{table*}

\begin{table*}[t!]
  
  \centering
  % \resizebox{\textwidth}{!}{%
  \begin{tabular}{@{}l|ccccc|ccccc@{}}
    \toprule
        Method & \multicolumn{5}{c|}{IOU$\uparrow$} & \multicolumn{5}{c}{Grad$\downarrow$} \\ 
        \multicolumn{1}{r|}{Shift} & \texttt{0} & \texttt{1} & \texttt{2} & \texttt{3} & \texttt{4} & \texttt{0} & \texttt{1} & \texttt{2} & \texttt{3} & \texttt{4} \\ 
        \midrule
        VADER (ours) & \bf 0.804 & \bf 0.801 & \bf 0.786 & \bf 0.760 & \bf 0.729 & \bf 0.0179 & \bf0.0180 & \bf0.0182 & \bf0.0189 & \bf0.0197 \\
        ICN \cite{ICN}  & 0.448 & 0.408 & 0.372 & 0.347 & 0.331  & 0.0245 & 0.0373 & 0.0408 & 0.0369 & 0.0343 \\
        ResNetConv~\cite{he2016deep} & 0.354 & 0.361 & 0.350 & 0.358 & 0.317  & 0.0241 & 0.0264 & 0.0226 & 0.0230 & 0.0242 \\
        SSD & 0.260 & 0.218 & 0.201 & 0.193 & 0.188  & 0.0230 & 0.0271 & 0.0226 & 0.0253 & 0.0276 \\
        % ErrAnalysis  \cite{ela} & 0.231 & 0.224 & 0.222 & 0.232 & 0.205  & 0.0309 & 0.0307 & 0.0302 & 0.0305 & 0.0331 \\
        % MantraNet \cite{mantranet2019cvpr} & 0.140 & 0.155 & 0.159 & 0.124 & 0.139  & 0.0698 & 0.0821 & 0.0637 & 0.0766 & 0.0917 \\
    \bottomrule
  \end{tabular}
  % }
  \caption{Differencing module evaluation with temporal jittering. Higher IOU scores between the predicted and ground truth heatmaps signify better accuracy of localization, lower values of gradients of the IOUs indicate more temporal stability. Both are reported for 5 different values of temporal shift, between 0 and 4.}
  \label{tab:iou_temp}
  \squeezeup
\end{table*}

The results in Table~\ref{tab:adachunk} indicate the importance of representing long videos with a variable number of temporal descriptors (as in Eq.~\ref{eq:ada_chunk}) to achieve good retrieval performance (comparing \textbf{All} to the other approaches). 
Furthermore, we observe consistently and significantly better performance from adaptive chunking compared to the constant chunking baselines at the same compression level.
Most noteworthy adaptive chunking can achieve the same or better retrieval performance at $2\times$ compression compared to the non-chunked baseline. 
This results in storage savings and faster response time in practice due to reduced index size. 
Given the comparable performance of consecutive chunk-similarity thresholding and clustering for adaptive chunking, we favor similarity thresholding in practice due to its simplicity and better control over the compression rate (the connection between the number of clusters and the compression rate is non-trivial).

\noindent \textbf{Aggregating Chunk-Level Search Results.}
We evaluate our proposed ranking and match verification score in Eq.~\ref{eq:vidscore1} and compare in Table~\ref{tab:scoring} to the following alternative formulations and baselines:
\noindent \textbf{Count} - We replace the score in Eq.~\ref{eq:vidscore1} with $\operatorname{score}(s)=\frac{1}{m}  \sum_{i=1}^{m} \mathbbm{1}[ s \in \{S_{i1}, \ldots, S_{ir} \} ] $, \ie, the proportion of query chunks for which a chunk of video $s$ was in the top $r$ retrievals.
\noindent \textbf{Max. Sim.} - We set $\operatorname{score}(s)$ to be the maximum similarity achieved for any retrieved chunk belonging to video $s$. 
\noindent \textbf{Uniform:} We replace the chunk weight in Eq.~\ref{eq:chunkweight} with $w_i(s)=1$ if $s \in \{S_{i1}, \ldots, S_{ir} \}$. The  score is an equal weighted sum of the best per-chunk similarities. 
% Unlike our formulation, it does not incorporate the rank or the variation in similarities of the retrievals. 
\noindent \textbf{Ours (w/o $\sigma$):} We set $\sigma_i=1$ in Eq.~\ref{eq:chunkweight} and remove dependence on per-query variation in similarities.

Table~\ref{tab:scoring} shows our scoring approach to improve the ranking of retrieved videos and significantly improves the F1 score vs. baselines. Our proposed weighting and aggregation of the chunk-level retrieval results thus can better identify correct matches, by incorporating the chunk-level ranking and variation of similarities in our method.

\noindent \textbf{Video Copy Detection on VCDB.}
We compare against prior approaches for video copy detection on VCDB \cite{jiang2014vcdb} in Table~\ref{tab:vcdb}. 
Our approach outperforms the prior works by a considerable margin and does so in a more scalable manner. 
The closest existing solution \cite{tan2022fast}, which also relies on an approximate nearest-neighbor lookup, still leverages additional networks to score and localize candidate matches.
In contrast, our approach solely parses the chunk-level retrieval results for coarse localization and scoring without incurring notable additional computing cost.

\subsection{Evaluating Alignment}
\label{sec:align}
We evaluate VADER's alignment performance by reporting the percentage of queries for which the alignment error is below a threshold $[0.1s, 1s, 10s]$.

We compare against three baselines -- for fair comparison we perform alignment using RMAC \cite{RMAC} features in all runs. 
We report the alignment performance results in Table \ref{tab:align} for four cases: \textbf{clean} - query clips are not edited, simply trimmed versions of the originals, \textbf{clean+benign} - simple in-place augmentations are applied to query clips, \textbf{manip} - clips are manipulated by professional video editors, \textbf{manip+benign} both of the above. 
In all cases, the query video $V_q$ is aligned with the original $V_o$. 
Table~\ref{tab:align} shows that our alignment method outperforms the other baselines in all four cases. Since the model is trained to be robust, the introduction of benign augmentations does not significantly affect  performance. While all methods perform worse on more challenging manipulations, VADER experiences a lower performance dip, compared to the closest baseline LAMV \cite{baraldi2018lamv}. 

% \begin{figure}[t!]
%     \centering
%     \includegraphics[width=1.0\linewidth]{figs/alignment.png}
%     \caption{Histogram of absolute alignment errors in number of frames (lower is better). For both the easiest (clean) and hardest (manip+benign) test sets. In both cases our proposed method has significantly more frames that are aligned exactly, with 0 error.}
%     \label{fig:align}
% \end{figure}

% \begin{figure}[t!]
%     \centering
%     \includegraphics[width=1.0\linewidth]{figs/retrieval_eg.png}
%     \caption{.}
%     \label{fig:ret}
% \end{figure}

% \begin{table}
%   \centering
%   \begin{adjustbox}{max width=\columnwidth}
%   \begin{tabular}{@{}lcccc@{}}
%     \toprule
%     Method | $\mathit{MSE}$ &  manip+benign & manip & benign & clear\\
%     \midrule
%     VADER (Ours) & \bf 173.0 & \bf 114.0 & \bf 22.3 & \bf 17.8 \\
%     NN & 184.5 & 163.8 & 31.5 & 28.9 \\
%     \bottomrule
%   \end{tabular}
%   \end{adjustbox}
%   \caption{Comparison of nearest neighbor (NN) and VADER video alignment approaches, showing mean squared error (MSE) between the predicted and ground truth frame ids.}
%   \label{tab:alignment}
% \end{table}

\subsection{Evaluating manipulation detection}
\label{sec:manip}
To generate the edited region outline, we upscale the $16\times 7 \times 7$ network output to the frame resolution $H\times W$, convert to it binary with a threshold and compute Intersection over Union (IoU) with the ground truth $\mathrm{IoU} = \frac{1}{16}\sum_{i=1}^{16}{\frac{U_i \cap T_i}{U_i \cup T_i}}$, where $T_i$ is the $H\times W$ binary ground truth heatmap, $U_i$ is the predicted heatmap after interpolation and thresholding. 

% The ground truth is initially already a binary mask of the frame, however it is downscaled to $16\times 7 \times 7$ grid format during model training for loss calculation.

We compare localization performance against five baselines. These baselines are image-based, and thus, we apply them to each frame pair separately. \textbf{Sum of Squared Distances (SSD)} - we compute SSD between two images at the pixel level, resize it to $7\times 7$, then resize it back before thresholding to create continuity in the detected heatmap.  \textbf{ResNetConv} - we extract $7\times 7\times2048$ features from pre-trained ImageNet ResNet50 model for both query and original images and calculate the distance between them. \textbf{ErrAnalysis} - inspired by the blind detection technique in \cite{ela}, we perform JPEG compression on the query and compare it with itself. \textbf{MantraNet} - is a supervised blind detection method \cite{mantranet2019cvpr} that detects anomalous regions. \textbf{Image Comparator Network (ICN)}\cite{ICN} - deep neural network for pairwise image comparison.

\begin{figure}[t!]
    \centering
    \includegraphics[width=1.0\linewidth,height=4cm]{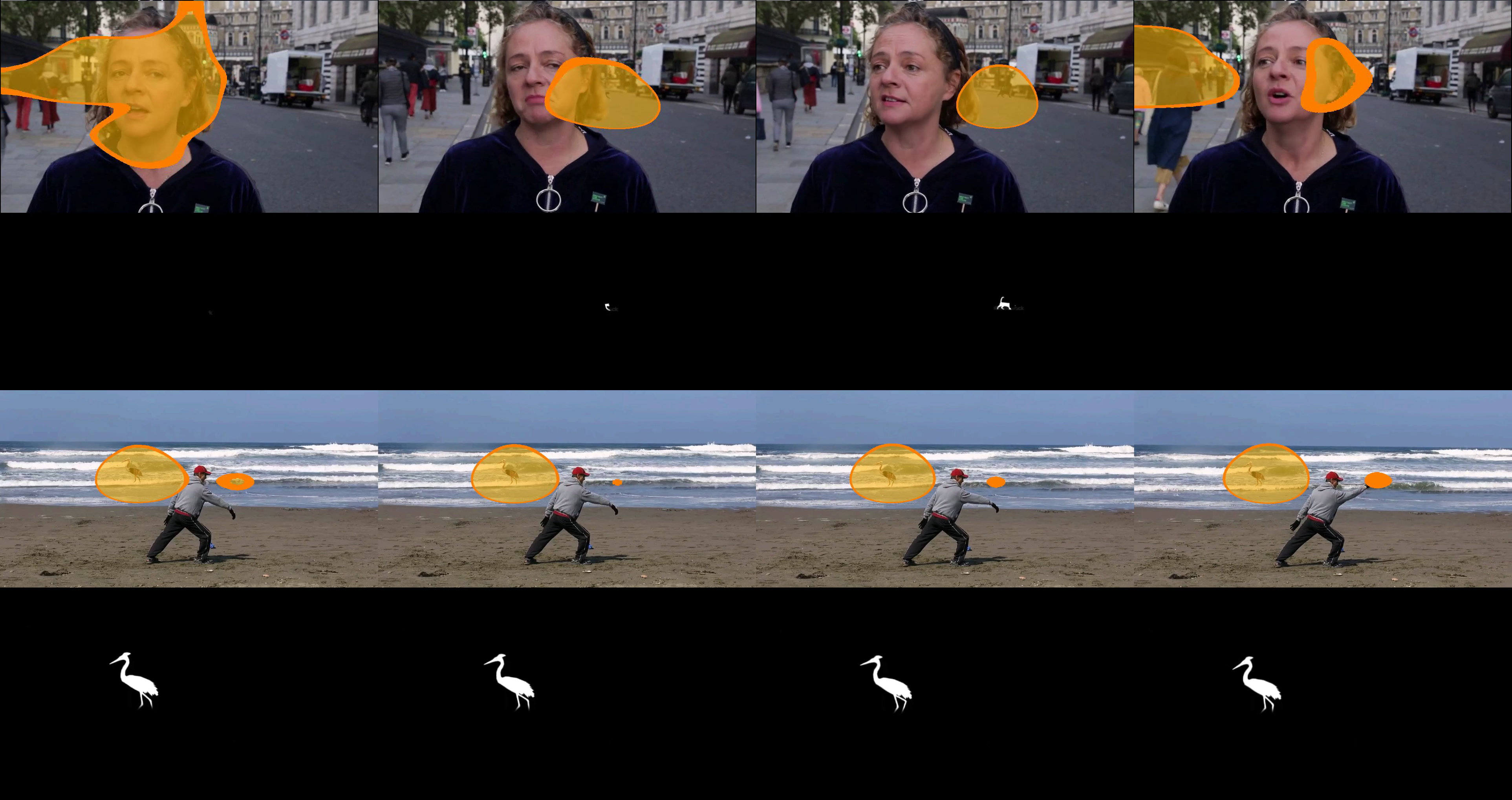}
    \caption{VADER limitations. Top: the model attempts to highlight the object inserted into the video even when it is occluded, resulting in some frames being falsely classified as manipulated. Bottom: cyclic actions of both the person and the background waves leads to incorrect alignment, which results in person's hand being incorrectly marked as manipulated.}
    \label{fig:fail}
\end{figure}

We introduce misalignment between two videos by offsetting all frames by a small amount (0-4 frames) and report IOU in Table~\ref{tab:iou_temp}. When the original and edited video are perfectly aligned, the image comparison methods are on par with scores expected when evaluating pairwise with frames. Our proposed method is able to leverage the temporal dimension of the data to achieve much stronger results in the presence of misalignment. To demonstrate robustness to benign transformations, in Table \ref{tab:iou_aug} we show how IOU changes with such transformation (see supmat for further evaluation of misalignment resilience). Our proposed space-time method exhibits strong temporal stability, lacking in pairwise frame based approaches. We quantify temporal stability as the mean gradient of IOU within all test videos. Table~\ref{tab:iou_temp} shows that even in the case of no misalignment, our method results in less flickering. Figure~\ref{fig:example} shows examples of the end-to-end VADER pipeline and we discuss limitations of alignment and differencing in Figure~\ref{fig:fail}.

\begin{table}[t!]
\centering
% \resizebox{\columnwidth}{!}{%
\begin{tabular}{lcc}
\hline
\multirow{2}{*}{\textbf{Method}} & \multicolumn{2}{c}{\textbf{IOU}} \\
 & \textbf{manip} & \textbf{manip + benign} \\ \hline
VADER & \textbf{0.804} & \textbf{0723} \\
ICN \cite{ICN} & 0.448 & 0.311 \\
ResNetConv \cite{he2016deep} & 0.354 & 0.174 \\
SSD & 0.260 & 0.205 \\
ErrAnalysis \cite{ela} & 0.231 & 0.058 \\
MantraNet \cite{mantranet2019cvpr} & 0.140 & 0.065 \\ \hline
\end{tabular}
% }

  \caption{Differencing module evaluation with in-place transformations. Higher IOU scores between the predicted and ground truth heatmaps signify better accuracy of localization.}
  \label{tab:iou_aug}
  \squeezeup
\end{table}

\section{Conclusion}

We presented VADER - a spatio-temporal method for robust video retrieval, alignment, and differencing.   VADER matches video fragments circulating online to a trusted database of original clips and associated provenance metadata. Providing users with information about video provenance and any  manipulations present  enables them to make  informed trust decisions about content they encounter, providing a solution to the 'soft binding' matching called for in emerging open standards for media provenance \eg C2PA \cite{c2pa}. Given a query video fragment, VADER retrieves the original and provides coarse temporal localization of the fragment within the original clip. VADER's alignment module produces a fine-grained frame-to-frame correspondence between the videos, refining the temporal localization. Aligned videos are compared via the space-time differencing module, which highlights the manipulations on each frame with a heatmap. The module is able to compensate for minor misalignments remaining through the end-to-end processing. We train and evaluate on ANAKIN - a novel dataset of manipulated videos and corresponding ground truth annotations.  The dataset is released as a further contribution of this paper.  Further work could address alternative ways to summarize manipulation in matched videos, for example captioning changes identified via the heatmap using a vision-language model.  This might enable larger numbers of results to be reviewed without high cognitive load.

\section*{Acknowledgement}
We thank the video editors who worked on ANAKIN: O. Tsiundyk\cite{upwork1}, A. Tsjornaja\cite{upwork2}, D.  Matheus\cite{upwork3}, N. Hassan\cite{upwork4}, I. Ievsiukov\cite{upwork5}, S. Skrypnik\cite{upwork6}.  Work  supported in part by DECaDE:  EPSRC Grant EP/T022485/1.

{\small
\bibliographystyle{ieee_fullname}
\bibliography{iccv23/main}
}

\end{document}